\documentclass{article}

\usepackage{microtype}
\usepackage{graphicx}
\usepackage{subfigure}
\usepackage{booktabs} 
\usepackage{multirow}
\usepackage{hyperref}
\usepackage{enumitem}

\usepackage[accepted]{icml2025}

\usepackage{amsmath}
\usepackage{amssymb}
\usepackage{mathtools}
\usepackage{amsthm}
\usepackage{graphicx}
\usepackage{subcaption}
\usepackage{colortbl}  
\usepackage{xcolor}    
\usepackage[capitalize,noabbrev]{cleveref}

\theoremstyle{plain}

\theoremstyle{definition}

\theoremstyle{remark}

\definecolor{lightgray}{RGB}{220, 220, 220}    
\definecolor{lightblue}{RGB}{198, 224, 244}    
\definecolor{lightgreen}{RGB}{197, 224, 180}   
\definecolor{lightpurple}{RGB}{230, 230, 250}  
\definecolor{lightorange}{RGB}{255, 218, 185}  
\usepackage[textsize=tiny]{todonotes}

\icmltitlerunning{Zero Token-Driven Deep Thinking in LLMs: Unlocking the Full Potential of Existing Parameters via Cyclic Refinement}

\begin{document}

\twocolumn[
\icmltitle{Zero Token-Driven Deep Thinking in LLMs: \\Unlocking the Full Potential of Existing Parameters via Cyclic Refinement}


\icmlsetsymbol{equal}{*}

\begin{icmlauthorlist}
\icmlauthor{Guanghao Li}{yyy,comp,sch}
\icmlauthor{Wenhao Jiang}{sch}
\icmlauthor{Li Shen}{sys}
\icmlauthor{Ming Tang}{comp}
\icmlauthor{Chun Yuan}{yyy}

\end{icmlauthorlist}

\icmlaffiliation{yyy}{SIGS, Tsinghua University}
\icmlaffiliation{comp}{Southern University of Science and Technology}
\icmlaffiliation{sch}{Guangdong Laboratory of Artificial Intelligence and Digital Economy (SZ)}
\icmlaffiliation{sys}{Sun Yat-sen University}
\icmlcorrespondingauthor{Wenhao Jiang}{cswhjiang@gmail.com}
\icmlcorrespondingauthor{Chun,Yuan}{yuanc@sz.tsinghua.edu.cn}

\icmlkeywords{Machine Learning, ICML}

\vskip 0.3in
]
\printAffiliationsAndNotice{}



\begin{abstract}
Resource limitations often constrain the parameter counts of Large Language Models (LLMs), hindering their performance. While existing methods employ parameter sharing to reuse the same parameter set under fixed budgets, such approaches typically force each layer to assume multiple roles with a predetermined number of iterations, restricting efficiency and adaptability.
In this work, we propose the {Zero Token Transformer (ZTT)}, which features a {head-tail decoupled parameter cycling} method. We disentangle the first (head) and last (tail) layers from parameter cycling and iteratively refine only the intermediate layers. Furthermore, we introduce a {Zero-Token Mechanism}, an internal architectural component rather than an input token, to guide layer-specific computation. At each cycle, the model retrieves a zero token (with trainable key values) from a Zero-Token Pool, integrating it alongside regular tokens in the attention mechanism. The corresponding attention scores not only reflect each layer’s computational importance but also enable dynamic early exits without sacrificing overall model accuracy.
Our approach achieves superior performance under tight parameter budgets, effectively reduces computational overhead via early exits, and can be readily applied to fine-tune existing pre-trained models for enhanced efficiency and adaptability.
\end{abstract}
    
\section{Introduction}
n recent years, it has been widely acknowledged that the performance of Large Language Models (LLMs) improves with an increasing number of parameters~\cite{rae2021scaling,rosenfeld2019constructive}. Consequently, scaling up parameter counts has become a common strategy for enhancing model performance~\cite{leviathan2023fast,xu2024fwdllm,pope2023efficiently}. However, this approach is often infeasible for users with limited computational resources. A critical challenge, therefore, is to achieve better performance under a \emph{fixed parameter budget}~\cite{zhou2024survey}.

A variety of model compression techniques, including quantization~\cite{lin2024awq,liu2023llm}, pruning~\cite{ma2023llm,sun2023simple}, and distillation~\cite{latif2023knowledge,shum2024first}, have been proposed to shrink large models to smaller ones. In parallel, another line of research has investigated ways to leverage additional computation within a fixed number of parameters, thereby unlocking deeper or more iterative reasoning~\cite{dehghani2018universal, lan2019albert}. A common strategy here is {parameter sharing}, where model layers reuse the same parameters across multiple computational cycles, sometimes referred to as “parameter cycling.” Rather than maintaining a separate set of parameters for each layer, models recurrently apply a compact parameter set, reducing memory requirements and potentially increasing depth of reasoning.

Despite its potential, parameter cycling raises three core challenges: (1) \textbf{Which} parameters should be reused across iterative cycles?
 (2) \textbf{How} can these shared parameters be managed to avoid functional conflicts and performance degradation?
(3) \textbf{When} should the model decide that no further reasoning is necessary, thus saving computational cost without truncating essential inference steps prematurely?

Existing works partially address one or two of these questions. For example, Solar~\cite{kim2023solar} reuses parameters from intermediate layers (\emph{which} parameters), while the Relaxed Recursive Transformer~\cite{bae2024relaxed} focuses on how to manage the recurring layer through LoRA~\cite{hu2021lora}. Palbert~\cite{balagansky2022palbert}, combining PonderNet~\cite{banino2021pondernet} with ALBERT, explores \emph{when} to stop via a dynamic pondering mechanism. However, none of these approaches provide a comprehensive solution that systematically addresses all three dimensions—\emph{which} parameters to cycle, \emph{how} to apply them, and \emph{when} to terminate reasoning.

In this paper, we propose a \emph{Zero Token Transformer (ZTT)} that systematically tackles these three challenges. Our approach is applicable to both training from scratch and fine-tuning existing pretrained models. Specifically:

\begin{itemize}
    \item \textbf{Head-Tail Decoupled Cyclic Architecture.} To handle the question of \emph{which} parameters to share, we decouple the first (head) and last (tail) layers from the parameter-sharing mechanism, because their specialized functions (encoding raw inputs and mapping representations to outputs) differ significantly from those of intermediate layers. Only the intermediate layers are recurrently used in a cyclic manner, improving efficiency while preserving essential input and output transformations.

    \item \textbf{Zero-Token Mechanism.} To address \emph{how} to manage shared parameters effectively, we introduce a novel {Zero-Token Mechanism}. Each intermediate layer retrieves a Zero-Token (with a trainable key and zero-valued representation) from a Zero-Token Pool and processes it alongside regular tokens. The attention scores toward the Zero-Token act as a guide for layer-specific computations, helping the model determine the extent of “reuse vs. new reasoning” at each cyclic iteration. This design mitigates potential conflicts that arise when reusing the same parameters multiple times.

    \item \textbf{A Dynamic Mechanism for Determining the Number of Cyclic Iterations.} Finally, to address \emph{when} to stop, we employ an early-exit style mechanism driven by the Zero-Token’s attention scores. When the attention to the Zero-Token surpasses a threshold, the model infers that further computation in subsequent cycles is unlikely to yield additional benefits and exits accordingly—striking a balance between computational efficiency and preserving accuracy.
\end{itemize}

The key contributions of this work can be summarized as follows:  
\begin{enumerate}  
    \item We propose a structured \emph{parameter cycling} approach that holistically tackles the “what, how, when” challenges of layer reuse under tight parameter budgets.
    
    \item We demonstrate that decoupling head and tail layers (while cycling among intermediate layers) yields both better reasoning depth and computational efficiency.
    
    \item We introduce the \emph{Zero-Token Mechanism}, allowing for dynamic control of layer-specific computation and enabling an effective early-exit strategy.
    
    \item 
    We demonstrate that our approach enhances performance in both training from scratch and fine-tuning existing large language models, highlighting its practical applicability to real-world deployments.

\end{enumerate}

\begin{figure*}[h]
    \centering
    \includegraphics[width=0.8\textwidth]{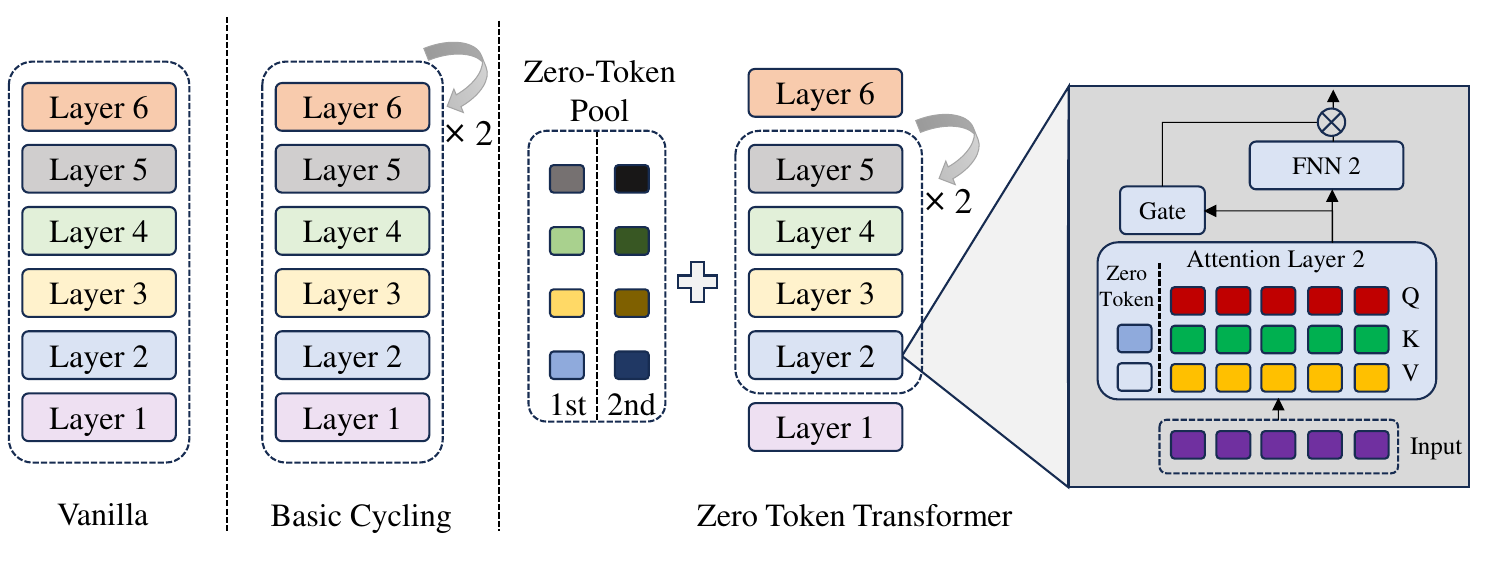}
    \vspace{-0.5cm}
    \caption{  
\textbf{Left:} A 6-layer vanilla transformer without cyclic processing.  
\textbf{Center:} A transformer with a simple two-cycle mechanism.  
\textbf{Right:} A two-cycle Zero Token  Transformer, where the first and last layers do not participate in the cycling process. Each layer introduces an additional Zero Token. The rightmost part illustrates how the Zero Token is incorporated. Using the second layer as an example: the Zero Token is prepended to the sequence by aligning its key with the original tokens at the beginning, and an all-zero value is added in front of the value sequence. Placing the Zero Token at the beginning ensures that all subsequent tokens can effectively attend to it.  
}

    \label{fig:your_image}
    \vspace{-0.5cm}
\end{figure*}

\section{Related Work}

Parameter sharing  has long been explored in early deep learning architectures, such as Convolutional Neural Networks (CNNs)~\cite{eigen2013understanding,savarese2019learning} and Recurrent Neural Networks (RNNs)~\cite{graves2016adaptive,sherstinsky2020fundamentals}, effectively reducing model complexity while preserving performance. The Universal Transformer~\cite{dehghani2018universal} later extended this idea to the Transformer architecture, demonstrating that reusing parameters in a cyclical manner across layers can substantially enhance efficiency. Subsequently, various studies have investigated \emph{which} Transformer components should be shared. Some focus on parameter reuse within individual layers~\cite{dabre2019recurrent}, tying encoder and decoder components~\cite{milbauer2023lait,xia2019tied}, or adopting partial expert networks~\cite{liu2024deepseek}. Others optimize \emph{how} parameter sharing is organized, for instance by stacking parameters in specific orders~\cite{takase2021lessons} or applying factorized embeddings~\cite{lan2019albert} to improve performance.
A critical aspect of parameter cycling is determining \emph{when} to repeat computations. Methods such as ACT~\cite{chowdhury2024recurrent,graves2016adaptive,csordas2024moeut,tan2023sparse} and PonderNet~\cite{banino2021pondernet,balagansky2022palbert} introduce adaptive recursion, allowing the model to decide how many cycles of computation are needed for deeper reasoning. However, most of these studies focus on training from scratch rather than fine-tuning large pre-trained models, limiting their practicality in real-world scenarios.

Recent work has begun to address parameter cycling within pre-trained Large Language Models. For instance, {Solar}~\cite{kim2023solar} improves performance by reusing parameters from the middle layers of the Llama model~\cite{touvron2023llama}, illustrating \emph{which} layers to cycle. Relaxed Recursive Transformers~\cite{bae2024relaxed} integrate LoRA~\cite{hu2021lora} modules to alleviate performance degradation caused by repetitive parameter usage. Despite these advances, the question of \emph{when} to stop recurrent processing remains underexplored. Although Relaxed Recursive Transformers consider varying numbers of cycles, they rely on fixed computation paths rather than a genuinely dynamic mechanism. Meanwhile, research on early exiting~\cite{chen2023ee,pan2024ee} focuses primarily on non-recurrent models, leaving open the question of how many cycles a recurrent model should undergo.
In contrast, our approach comprehensively addresses the three central questions of \emph{what} to cycle, \emph{how} to manage recurrent parameters, and \emph{when} to terminate reasoning. We propose a method applicable both to training from scratch and to fine-tuning pre-trained LLMs, offering a practical and efficient solution for enhancing model performance under tight parameter budgets.

\begin{figure*}[h]
    \centering
    \includegraphics[width=0.9\textwidth]{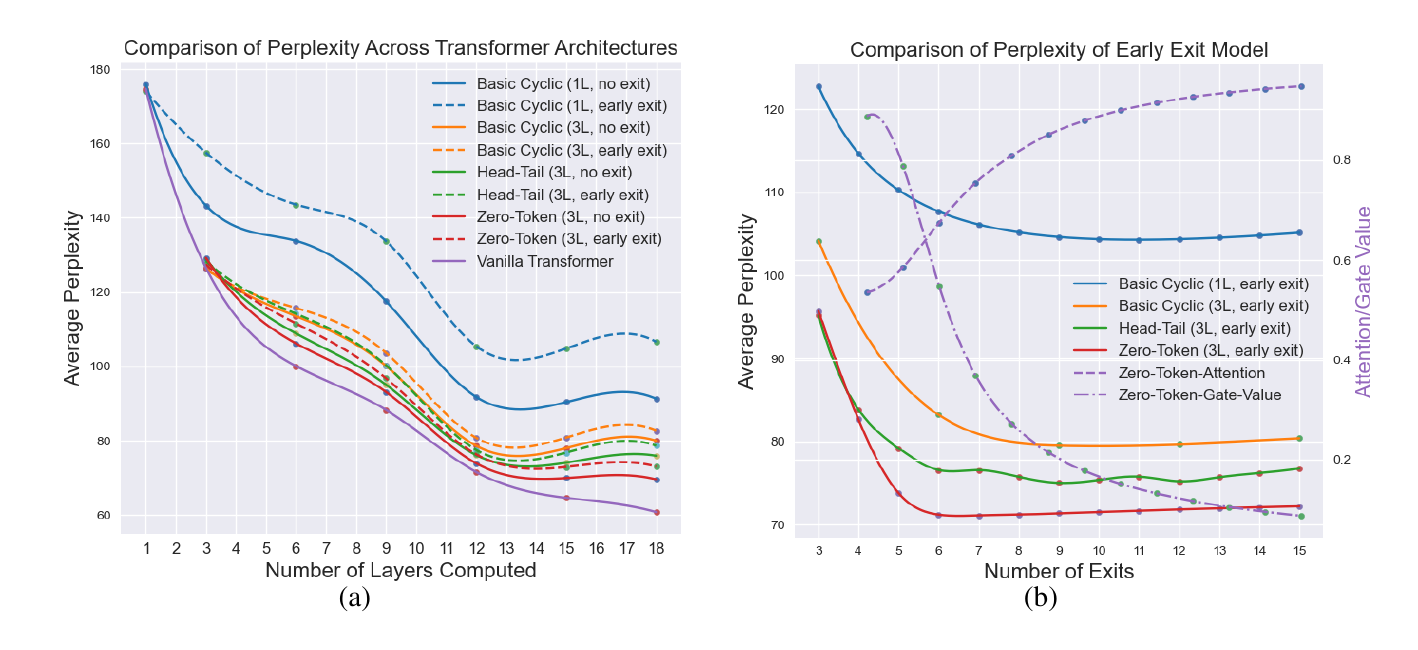}
    \vspace{-0.5cm}
\caption{
Comparison of model performance under equal computational complexity.
\textbf{(a)} The effect of varying computational complexity, where 1L denotes the original model with a single layer, and increased complexity corresponds to repeated model calls.
``Early exit'' refers to adding a classification head after each cycle to train intermediate results.
\textbf{(b)} On the left y-axis, the intermediate results of different models under the ``early exit'' condition when the total computational complexity is fixed at 15 (cycles $\times$ layers).
On the right y-axis, the average attention values of other tokens to the Zero Token and the gate value at the output of the Zero Token Transformer.
}
\label{fig:1}
\vspace{-0.5cm}
\end{figure*}

\section{Zero Token Transformer}
In this section, we introduce our \emph{Zero-Token Transformer (ZTT)}, a novel approach that combines \emph{Head-Tail Decoupled Cyclic Architecture} and a \emph{Zero-Token Mechanism} with a gating strategy.
We begin by examining preliminary observations from a \emph{Basic Cyclic Transformer (BCT)} and highlight the challenges that arise when scaling up.
We then describe how our \emph{head-tail decoupling} addresses these issues, and how the \emph{Zero Token} and \emph{gating mechanism} further enhance performance and enable dynamic early exiting.

\subsection{Basic Cyclic Transformer (BCT) and Motivation}
\label{sec:bct}

We first consider a simple form of cyclic Transformer, following~\cite{bae2024relaxed,takase2021lessons}, where a single Transformer layer (or a small stack of layers) is repeatedly applied for $N$ cycles.
Formally, let $H_F^{l,n}$ be the output of layer $l$ at cycle $n$:

\vspace{-0.5cm}
\begin{equation}
\label{eq:1}
\begin{aligned}  
    H_F^{l,n} &=  H_A^{l,n} + \text{FFN}\bigl(\text{LN}(H_A^{l,n}), \theta^l\bigr),\\
    H_A^{l,n} &= H_F^{l-1,n} + \text{MultiHead}\bigl(H_F^{l-1,n}, \Phi\bigr),\\
    l &\in \{1, 2, \dots, L\}, \quad n \in \{1, 2, \dots, N\},
\end{aligned}
\end{equation}
\vspace{-0.5cm}

where $\Phi$ and $\theta^l$ are trainable parameters, and the same parameters are reused at each cycle. For simplicity, $H_F^{0,1}$ is the embedding of the input tokens. When $N=1$, this reduces to a standard (vanilla) $L$-layer Transformer.

Figure~\ref{fig:1}(a) shows perplexities on WikiText-2 under the same total computational cost (\emph{number of layers} $\times$ \emph{number of cycles}). 
For instance, a 1-layer BCT run for 12 cycles achieves a lower perplexity than a 1-layer vanilla Transformer (i.e., $N=1$), but it is still worse than a standard 12-layer Transformer with the same total computational budget.
Similarly, adding classification heads after each cycle for early exiting (``Early exit'' in Figure~\ref{fig:1}(a)) further degrades performance, indicating that requiring intermediate outputs imposes extra burdens on the cyclic layers.

\paragraph{Empirical Comparison.}
These findings highlight a fundamental issue with basic cyclic Transformers:
each \emph{cyclic block} assumes significantly greater responsibilities than a vanilla Transformer layer. 
First, cyclic layers must fulfill the roles of multiple layers by providing enhanced intermediate representations for subsequent computations, while also producing high-quality outputs for the classification head.
Moreover, there is no explicit guidance on the specific role a given layer should perform at each iteration. 
This lack of role distinction may lead to confusion and computational conflicts, ultimately affecting overall performance. 
Consequently, even when the current representation is well-optimized, it is recalculated in subsequent cycles, potentially overwriting previously learned representations.

\paragraph{Increasing Cycles Alone Is Insufficient.}
Reusing the same layer(s) for multiple cycles indeed improves performance compared to a single-layer Transformer.
However, as we increase the total layer count or require early-exit outputs at each cycle, the \emph{performance gap} between Basic Cyclic Transformers and an equivalently sized \emph{vanilla} Transformer \emph{widens}.
We hypothesize that this occurs because each cyclic block in BCT must simultaneously compute refined internal representations \emph{and} produce final outputs for classification at each cycle, without any guidance on how to specialize. 
As cycles accumulate, these conflicting objectives can lead to overwriting or “conflicting” representations.

Based on the observations above, we identify three key \emph{issues} that motivate our design:
\begin{enumerate}[label=\textbf{Issue \arabic*:}, leftmargin=1.2cm, itemsep=2pt, topsep=2pt]
    \item \textbf{No separation of specialized layers.} 
    The first and last layers in a Transformer typically have distinct roles (e.g., mapping raw inputs or producing logits). Forcing them to share parameters can degrade performance (\S\ref{sec:hdt}).

    \item \textbf{Lack of role distinction among cycles.} 
    In BCT, the same layer repeatedly processes the representation, even if the current representation is sufficiently refined. There is no mechanism to skip certain cycles or to mark a cycle as “for further refinement” (\S\ref{sec:ztm}).

    \item \textbf{When to stop further computation?}
    Simply running $N$ cycles can waste computation once the network is “confident.” Likewise, forcing a classification output at \emph{every} cycle can degrade performance. A more \emph{dynamic} approach is needed (\S\ref{sec:gating}).
\end{enumerate}

To address these issues, we propose the \emph{Zero-Token Transformer}, which comprises:

- \textbf{Head-Tail Decoupled Cyclic Architecture} (\S\ref{sec:hdt}): We do not cycle the first and last layers, preserving their specialized roles while only reusing \emph{intermediate} layers.
- \textbf{Zero-Token Mechanism} (\S\ref{sec:ztm}): We insert a learnable “Zero Token” into each attention layer to guide or skip computations dynamically, enabling distinct cycle-specific roles.
- \textbf{A Dynamic Mechanism for Determining the Number of Cyclic Iterations} (\S\ref{sec:gating}): We add a lightweight gating network in the feed-forward layer to help decide when to terminate further computation based on the Zero Token's attention.

In the following subsections, we detail each component and explain how they address the issues above.

\subsection{Head-Tail Decoupled Cyclic Architecture}
\label{sec:hdt}

Recent analyses~\cite{sun2024transformer} suggest that \emph{intermediate} Transformer layers often exhibit functionally similar representations, whereas the \emph{head} (first) and \emph{tail} (last) layers specialize in tasks such as raw feature encoding and output mapping. 
Pruning or drastically altering these boundary layers tends to have an outsized impact on overall performance.

Hence, we preserve the original first and last layers as \emph{fixed} (non-cyclic) and let only the middle layers reuse parameters across $N$ cycles, as shown in Figure~\ref{fig:your_image}(right).
Concretely, if the Transformer has $L$ layers, we exclude layers $1$ and $L$ from parameter cycling:
\vspace{-0.1cm}
\begin{equation}
l \in \{2,3,\dots,L-1\}, \quad n \in \{1,\dots,N\}.
\end{equation}
By doing so, the distinct responsibilities of the head and tail layers remain intact, mitigating conflicts. Figure~\ref{fig:1}(a) shows that under the same total computational budget, our Head-Tail Decoupled Cyclic Transformer (HDT) achieves better perplexity than a straightforward “all-layer cycling” design, confirming that preserving specialized boundary layers helps alleviate the performance gap (Issue 1).

\subsection{Zero-Token Mechanism}
\label{sec:ztm}

Even with head-tail decoupling, intermediate layers in a cyclic setup can still \emph{redundantly} process representations. 
If the current representation is already high-quality, repeatedly refining it may overwrite earlier features or create conflicts. 
We address this by introducing a small, trainable “Zero Token” into each attention layer—acting like a prompt that “signals” whether the model should refine or skip a cycle.

For the $l$-th layer in cycle $n$, we insert a Zero Token (ZToken) at the start of the sequence. It has:
\begin{itemize}[leftmargin=1.1em]
    \item A \textbf{key} vector $K_{z,i}^{l,n}$ (split by head $i$) that is \emph{trainable},
    \item A \textbf{value} vector $V_{0,i}^{l,n}$ that is \emph{all zeros}, and
    \item \textbf{No query} component.
\end{itemize}
Placing the Zero Token at the front ensures all other tokens can attend to it. Formally, the multi-head attention in Eq.~\ref{eq:1} becomes:
\begin{equation}
\label{eq:kv}
\begin{aligned}
K_{\text{new}, i}^{l,n} &= \bigl[K_{z,i}^{l,n},\, K_i^l\bigr], \\
V_{\text{new}, i}^{l,n} &= \bigl[V_{0,i}^{l,n},\, V_i^l\bigr], 
\end{aligned}
\quad
\text{MultiHead}\bigl(Q^l, K_{\text{new}}^{l,n}, V_{\text{new}}^{l,n}\bigr).
\end{equation}

\paragraph{Role of the Zero Token.}
Each cycle “fetches” its own Zero Token, whose trainable key can induce high or low attention from the queries $Q^l$. 
If the model “pays a lot of attention” to this zero-valued token, the output effectively becomes the same as the previous representation (since multiplying by zero yields no further update). 
This lets each layer decide whether to \emph{refine} or \emph{skip}, addressing \emph{Issue 2} (lack of role distinction across cycles).

In Figure~\ref{fig:1}(b) (right y-axis), we plot the average attention score to the Zero Token over different cycles. 
Higher attention corresponds to the model “opting out” of deeper recalculation.

\subsection{A Dynamic Mechanism for Determining the Number of Cyclic Iterations}
\label{sec:gating}

While the Zero Token's attention score is the main indicator for deciding whether to halt additional cycles, we introduce a lightweight gating mechanism around the feed-forward network (FFN) to provide finer-grained computational control. 
Even if the model has not yet triggered an early exit, some cycles may only require partial FFN computation.

We modify the FFN in Eq.~\ref{eq:1} by adding:
\begin{equation}
\label{eq:gating}
\begin{aligned}
H_F^l &= H_A^l 
+ \Bigl[\text{FFN}\bigl(\text{LN}(H_A^l), \theta^l\bigr)\Bigr] 
  \cdot \text{gate}\bigl(\text{LN}(H_A^l)\bigr),\\
\text{gate}(\cdot) &\in [0,1].
\end{aligned}
\end{equation}
When $\text{gate}(\cdot)$ is close to 1, the full FFN transform is applied; when near 0, the FFN is mostly bypassed, saving computation. 
We observe that the gating value often correlates with the Zero Token’s attention: if the model pays high attention to the Zero Token, it implies less refinement is needed, and the gate decreases accordingly.

\paragraph{Early Exit Criterion.}
The Zero Token’s attention score (\S\ref{sec:ztm}) remains the primary criterion for early exit. 
Once it exceeds a threshold (e.g., 0.9), we terminate further cycles and output the final representation. 
The gating function simply provides a smoother transition for intermediate cycles, reducing unnecessary FFN computation \emph{before} the final exit trigger.

\paragraph{Discussion.}
We do not gate the \emph{attention module} itself because attention integrates token representations, including the Zero Token. 
By gating the FFN, we allow partial skipping of the more expensive transformations without interfering with the Zero Token’s signaling. 
Hence, the Zero Token attention decides \emph{when} to stop entirely, while the gate refines \emph{how much} computation to apply until that point.

Overall, by combining head-tail decoupling (to preserve specialized boundary layers), a Zero Token (to guide cycle-level computation), and a gating-based mechanism (to smooth partial skips and enable early exit), our Zero-Token Transformer effectively addresses the three major issues outlined in \S\ref{sec:bct}.

\section{Experiments}
\label{sec:experiments}
\renewcommand{\arraystretch}{1.1}
\begin{table*}[]
\small  
\caption{ 
Evaluation results of different models pre-trained on the C4 dataset and fine-tuned on the test datasets, including PIQA, ARC-Challenge, ARC-Easy, LAMBADA, and HellaSwag. We report accuracy for each dataset. The computation formula used for each model in the table is represented as: \textit{All Layers - Looped Layers + (Looped Layers $\times$ Loop Count)}.
}
\label{tab:scartch}
\begin{tabular}{c|c|c|c|c|ccccc|c|c}
\hline
\rowcolor{lightgray}
\textbf{Models}                                                                        & \textbf{Size}           & \textbf{\begin{tabular}[c]{@{}c@{}}All \\ Layers\end{tabular}} & \textbf{\begin{tabular}[c]{@{}c@{}}Looped \\ Layers\end{tabular}} & \textbf{\begin{tabular}[c]{@{}c@{}}Loop \\ Count\end{tabular}} & \textbf{PQ} & \textbf{ARC-c} & \textbf{ARC-e} & \textbf{LD} & \textbf{HS} & \textbf{Avg} & \textbf{Model\_Avg}    \\ \hline
V\_small                                                                         & 60.65M                  & 3                                                              & 0                                                                 & -                                                              & 61.32       & 17.83          & 37.84          & 13.8        & 26.75       & 31.51        & 31.51                  \\ \hline
V                                                                                & 81.9M                   & 6                                                              & 0                                                                 & -                                                              & 64.15       & 19.28          & 40.32          & 19.08       & 27.04       & 33.97        & 33.97                  \\ \hline
\rowcolor{lightpurple}
BC                                                                         & 60.65M                  & 3                                                              & 3                                                                 & -                                                              & 63.44       & 18.6           & 38.8           & 16.32       & 26.77       & 32.79        & 32.79                  \\ \hline
\multirow{2}{*}{\begin{tabular}[c]{@{}c@{}}BCE\end{tabular}} & \multirow{2}{*}{60.65M} & \multirow{2}{*}{3}                                             & \multirow{2}{*}{3}                                                & 1                                                              & 61.7        & 17.83          & 37.88          & 13.24       & 26.93       & 31.52        & \multirow{2}{*}{32.17} \\
                                                                                       &                         &                                                                &                                                                   & 2                                                              & 62.73       & 18.17          & 39.94          & 16.32       & 26.9        & 32.81        &                        \\ \hline
\rowcolor{lightgreen}
HTC                                                                              & 60.65M                  & 3                                                              & 1                                                                 & 4                                                              & 63.17       & 19.2           & 40.15          & 16.52       & 26.73       & 33.15        & 33.15                  \\ \hline
\multirow{4}{*}{\begin{tabular}[c]{@{}c@{}}HTCE\end{tabular}}      & \multirow{4}{*}{60.65M} & \multirow{4}{*}{3}                                             & \multirow{4}{*}{1}                                                & 1                                                              & 63.55       & 18.77          & 39.65          & 16.24       & 26.67       & 32.98        & \multirow{4}{*}{32.62} \\
                                                                                       &                         &                                                                &                                                                   & 2                                                              & 64.04       & 17.92          & 40.87          & 13.33       & 26.55       & 32.54        &                        \\
                                                                                       &                         &                                                                &                                                                   & 3                                                              & 62.95       & 18.26          & 39.73          & 15.33       & 26.68       & 32.59        &                        \\
                                                                                       &                         &                                                                &                                                                   & 4                                                              & 63.71       & 18.69          & 39.86          & 12.78       & 26.79       & 32.37        &                        \\ \hline
\rowcolor{lightorange}
ZTT                                                                            & 61.77M                  & 3                                                              & 1                                                                 & 4                                                              & 62.51       & 19.3           & 40.87          & 17.94       & 26.98       & 33.52        & 33.52                  \\ \hline
\multirow{4}{*}{\begin{tabular}[c]{@{}c@{}}ZTTE\end{tabular}}     & \multirow{4}{*}{61.77M} & \multirow{4}{*}{3}                                             & \multirow{4}{*}{1}                                                & 1                                                              & 62.95       & 17.66          & 38.76          & 16.71       & 26.76       & 32.57        & \multirow{4}{*}{32.79} \\
                                                                                       &                         &                                                                &                                                                   & 2                                                              & 63.44       & 18.26          & 41.16          & 16.63       & 26.81       & 33.26        &                        \\
                                                                                       &                         &                                                                &                                                                   & 3                                                              & 64.25       & 17.58          & 40.61          & 14.28       & 26.85       & 32.71        &                        \\
                                                                                       &                         &                                                                &                                                                   & 4                                                              & 63.55       & 18             & 41.04          & 13.59       & 26.93       & 32.62        &                        \\ \hline
\end{tabular}
 \vspace{-0.5cm}
\end{table*}  
In this section, we present our experimental setup and results to demonstrate the effectiveness of our proposed \textit{Zero-Token Transformer} approach under a fixed parameter budget. We evaluate both {training from scratch} and {fine-tuning} scenarios, using a decoder-only Transformer architecture.

\subsection{Experimental Setup}
\label{sec:exp_setup}

\textbf{Models.}
We consider two main training settings:
\begin{itemize}[leftmargin=*]
    \item \textbf{Training from Scratch:} We base our architecture on GPT-2~\cite{radford2019language} but restrict each layer to around 10M parameters. The total number of layers is $L=6$. To maintain a fixed computational budget, we define a total of 6 ``network computation cycles'',  where each layer in a standard setting (i.e., without parameter sharing) is counted as one cycle. All models in this setting are pre-trained on the C4 English subset~\cite{raffel2020exploring} using a causal next-token prediction objective for 10B tokens.
    \item \textbf{Fine-Tuning Pre-Trained Models:} We also fine-tune widely used checkpoints such as {GPT-2} and {OPT}~\cite{zhang2023opt} to show that our approach can be applied to large pre-trained models with minimal modification.
\end{itemize}

\textbf{Baselines.}
We compare several approaches, all based on decoder-only Transformers. One variant, referred to as early exit, adds classification heads at each cycle, facilitating intermediate predictions. These intermediate exits are represented as E.

\begin{itemize}[leftmargin=*]
    \item \textbf{Vanilla (V):} A standard Transformer with $L$ distinct layers. The total computation cost is effectively $L$ cycles.
    \item \textbf{Basic Cycling (BC):} The model has $L$ layers but shares parameters across layers by cycling them $N$ times. This results in a total computation cost of $L \times N$. 
    \item \textbf{Head-Tail Cycling (HTC):} Instead of cycling \emph{all} $L$ layers, only the \emph{intermediate} layers are reused $N$ times, while the first and last layers remain distinct. This structure aims to preserve the special functions of the input and output layers. We also consider an early exit variant of {this scheme.}
    \item \textbf{Zero-Token Transformer (ZTT):} Our proposed method, which only cycles the intermediate layers and introduces a \emph{Zero Token} in each attention layer. This token provides a learnable \emph{key} (prompt-like) while carrying zero-value vectors, enabling the model to distinguish between different cycles and facilitate \emph{adaptive computation}. {We also include an {early exit} variant } that uses the learned \emph{Zero Attention} signals to decide when to stop.
\end{itemize}

\textbf{Evaluation Datasets and Metrics.}
We considered nine different types of datasets: 
(a) \textit{Reasoning}: PIQA~\cite{bisk2020piqa}; 
(b) \textit{Multiple Choice}: ARC Challenge~\cite{clark2018think}, ARC Easy~\cite{clark2018think}; 
(c) \textit{Long-Term Context Recall}: LAMBADA~\cite{paperno2016lambada} and 
(d) \textit{Natural Language Inference}: HellaSwag~\cite{zellers2019hellaswag}.

For multiple-choice tasks, we report {accuracy}, and for LAMBADA we report {exact match} accuracy on the held-out set. We employ the Language Model Evaluation Harness~\cite{eval-harness} for consistent evaluations. All models {pre-trained from scratch} are fine-tuned on each downstream task before testing. Similarly, the large pre-trained checkpoints (GPT-2, OPT) are directly fine-tuned on these tasks using the same hyperparameter settings (details in ~\cref{es}).

\vspace{-0.2em}

\subsection{Results of Training from Scratch}
\label{sec:exp_scratch}

Table~\ref{tab:scartch}summarizes the performance of models trained {from scratch} under a fixed parameter budget. We highlight the following observations:

\textbf{Effect of Fewer Layers (Vanilla vs.\ Vanilla-Small).}
When we reduce both the parameter count and computational budget from $L=6$ to $L=3$ (``Vanilla-small''), the accuracy drops significantly (e.g., from 33.97\% to 31.51\%). {This indicates that simply using fewer layers cannot maintain adequate performance without cycling.}

\textbf{Basic Cycling (BC).}
To mitigate the performance gap, {BC} reuses a 3-layer stack twice (for a total of 6 cycles), partially recovering performance to 32.79\%. This confirms that increased ``computational depth'' via cycling can help, though it still lags behind the original 6-layer Transformer. Introducing {early exit} ({BCE}) in BC leads to a slight accuracy drop (32.17\%), suggesting that training additional intermediate heads can sometimes introduce optimization trade-offs.

\textbf{Head-Tail Cycling (HTC).}
By fixing the first and last layers while only cycling the intermediate ones, {we achieve 33.15\% (no early exit) and 32.62\% (early exit), surpassing Basic Cycling.} This underscores the importance of preserving specialized head and tail layers.

\textbf{Zero-Token Transformer (ZTT).}
Building on head-tail separation, our proposed {Zero Token} mechanism further boosts accuracy to 33.52\% without early exit, and 32.79\% with early exit. Notably, ZTT consistently outperforms both BC and HT across tasks, demonstrating that the Zero-Token mechanism effectively guides each cycle’s computation and alleviates functional conflicts in shared parameters.

Overall, these results confirm the benefit of our {ZTT} approach in balancing parameter efficiency and modeling capacity. Even when the total parameter budget and computation cycles are constrained, introducing Zero Tokens with a head-tail separation strategy yields superior accuracy.

\subsection{Adaptive Inference with Zero Attention}
\label{sec:adaptive_inference}

\renewcommand{\arraystretch}{1.1}
\begin{table}[]
\centering
\small  
\caption{
Perplexity (PPL) and Zero Attention metrics of the Zero Token Transformer (early exit) method on the C4 dataset with different loop counts. Zero Attention refers to the average attention of other tokens to the Zero Token, while Gate Value represents the output of the gating unit in the model's FNN layer.
}
\vspace{-0.4cm}
\label{tab:attention}
\begin{tabular}{c|cccc}
\hline
 
\textbf{Loop Count} & \textbf{1} & \textbf{2} & \textbf{3} & \textbf{4} \\ \hline
Zero Attention      & 0.21       & 0.47       & 0.54       & 0.65       \\ \hline
Gate Value          & 0.55       & 0.32       & 0.20       & 0.15       \\ \hline
PPL                 & 37.02      & 34.58      & 34.03      & 33.99      \\ \hline
\end{tabular}
 \vspace{-0.96cm}
\end{table}

We next investigate whether \textit{Zero Attention}—the average attention placed on the Zero Token—can serve as a stopping criterion for adaptive inference. Specifically, once the Zero Attention score surpasses a predefined threshold \( P \), we consider the representation sufficiently refined and terminate further cycles.

In Table~\ref{tab:attention}, we track perplexity ({PPL}), Zero Attention, and Gate Value across different cycle counts. As the loop depth increases, {Zero Attention} gradually rises, while the {Gate Value} in the feed-forward network decreases, suggesting diminishing returns from additional computation.

Table~\ref{tab:adaptive} further examines how different thresholds \( P \) affect both model accuracy and the average number of computation cycles. A lower threshold (\( P = 0.2 \)) forces early termination, significantly reducing compute but at the cost of some performance loss. In contrast, a moderate threshold (\( P = 0.5 \)) provides a strong balance, achieving 33.00\% accuracy with an average of just 3.31 cycles—matching or even surpassing the full 4-cycle baseline. Increasing the threshold further (\( P = 0.7 \)) results in more reasoning steps, leading to slight accuracy improvements but at the expense of higher computational costs.

These findings illustrate that Zero Attention can effectively guide {dynamic computation}, allowing models to adaptively allocate reasoning cycles while maintaining strong performance. This presents a promising strategy for efficient inference in resource-constrained settings.

\begin{table}[]
\centering
\small  
\caption{
A table showing the selection of different threshold values \( P \) for the average attention of tokens in the attention mechanism as a criterion for controlling when the model stops reasoning. The table presents the model's accuracy on different tasks under various threshold values, as well as the average number of cycles required for the looping layers.
}

\label{tab:adaptive}
\begin{tabular}{c|cccc}
\hline

\textbf{P} & \textbf{0.2} & \textbf{0.5} & \textbf{0.7} & 1     \\ \hline
Avg\_Acc   & 32.42        & 33        & 33.08        & 32.62 \\ \hline
Avg\_Loop  & 1.58         & 3.31         & 3.73         & 4     \\ \hline
\end{tabular}
 \vspace{-0.8cm}
\end{table}

\subsection{Fine-Tuning Results on Pre-Trained Models}
\label{sec:exp_pretrained}

\renewcommand{\arraystretch}{1.2}
\begin{table*}[ht]
\centering
\small  
\caption{
Results of different fine-tuned pre-trained models on multiple tasks. The abbreviations used in the table are: Vanilla (V), Simple Cycling (BC), Simple Cycling with early exit (BCE), Zero Token Transformer (ZTT), and Zero Token Transformer with early exit (ZTTE).
}
\label{tab:pertrain}
\begin{tabular}{cc|c|c|c|c|ccccc|c|c}
\hline

\multicolumn{2}{c|}{Models}                                                                                                                   & \textbf{Size}                                     & \textbf{\begin{tabular}[c]{@{}c@{}}All \\ Layers\end{tabular}} & \textbf{\begin{tabular}[c]{@{}c@{}}Looped \\ Layers\end{tabular}} & \textbf{\begin{tabular}[c]{@{}c@{}}Loop \\ Count\end{tabular}} & \textbf{PQ}                   & \textbf{ARC-c}                & \textbf{ARC-e}                & \textbf{LD}                   & \textbf{HS}                   & \textbf{Avg}                  & \textbf{Model\_Avg}                                      \\ \hline
\multicolumn{1}{c|}{}                                                                         & V                                             & 125.24M                                           & 12                                                             & 0                                                                 & -                                                              & 62.89                         & 19.03                         & 43.52                         & 28.95                         & 29.19                         & 36.72                         & 36.72                                                    \\ \cline{2-13} 
\multicolumn{1}{c|}{}                                                                         & BC                                            & 125.24M                                           & 12                                                             & 12                                                                & 2                                                              & 65.67                         & 21.33                         & 43.39                         & 38.44                         & 28.71                         & 39.51                         & 39.51                                                    \\ \cline{2-13} 
\multicolumn{1}{c|}{}                                                                         &                                               &                                                   &                                                                &                                                                   & 1                                                              & 65.02                         & 20.05                         & 43.8                          & 35.2                          & 28.83                         & 38.58                         &                                                          \\
\multicolumn{1}{c|}{}                                                                         & \multirow{-2}{*}{BCE}                         & \multirow{-2}{*}{125.24M}                         & \multirow{-2}{*}{12}                                           & \multirow{-2}{*}{12}                                              & 2                                                              & 65.67                         & 21.25                         & 43.94                         & 31.86                         & 28.97                         & 38.34                         & \multirow{-2}{*}{38.46}                                  \\ \cline{2-13} 
\multicolumn{1}{c|}{}                                                                         & ZTT                                            & 129.68M                                           & 12                                                             & 10                                                                & 2                                                              & 65.29                         & 21.73                         & 44.02                         & 38.51                         & 28.9                          & 39.69                         & 39.69                                                    \\ \cline{2-13} 
\multicolumn{1}{c|}{}                                                                         & \cellcolor[HTML]{C0C0C0}                      & \cellcolor[HTML]{C0C0C0}                          & \cellcolor[HTML]{C0C0C0}                                       & \cellcolor[HTML]{C0C0C0}                                          & \cellcolor[HTML]{C0C0C0}1                                      & \cellcolor[HTML]{C0C0C0}66    & \cellcolor[HTML]{C0C0C0}20.65 & \cellcolor[HTML]{C0C0C0}43.43 & \cellcolor[HTML]{C0C0C0}33.57 & \cellcolor[HTML]{C0C0C0}28.8  & \cellcolor[HTML]{C0C0C0}38.49 & \cellcolor[HTML]{C0C0C0}                                 \\
\multicolumn{1}{c|}{\multirow{-7}{*}{OPT}}                                                    & \multirow{-2}{*}{\cellcolor[HTML]{C0C0C0}ZTTE} & \multirow{-2}{*}{\cellcolor[HTML]{C0C0C0}129.68M} & \multirow{-2}{*}{\cellcolor[HTML]{C0C0C0}12}                   & \multirow{-2}{*}{\cellcolor[HTML]{C0C0C0}10}                      & \cellcolor[HTML]{C0C0C0}2                                      & \cellcolor[HTML]{C0C0C0}66.16 & \cellcolor[HTML]{C0C0C0}20.9  & \cellcolor[HTML]{C0C0C0}42.8  & \cellcolor[HTML]{C0C0C0}33.96 & \cellcolor[HTML]{C0C0C0}28.82 & \cellcolor[HTML]{C0C0C0}38.53 & \multirow{-2}{*}{\cellcolor[HTML]{C0C0C0}\textbf{38.51}} \\ \hline
\multicolumn{1}{c|}{}                                                                         & V                                             & 124.44M                                           & 12                                                             & 0                                                                 & -                                                              & 62.89                         & 19.03                         & 43.81                         & 25.97                         & 28.92                         & 36.12                         & 36.12                                                    \\ \cline{2-13} 
\multicolumn{1}{c|}{}                                                                         & BC                                            & 124.44M                                           & 12                                                             & 12                                                                & 2                                                              & 65.23                         & 20.65                         & 43.35                         & 29.34                         & 28.26                         & 37.37                         & 37.37                                                    \\ \cline{2-13} 
\multicolumn{1}{c|}{}                                                                         &                                               &                                                   &                                                                &                                                                   & 1                                                              & 64.69                         & 20.22                         & 43.81                         & 30.22                         & 28.39                         & 37.47                         &                                                          \\
\multicolumn{1}{c|}{}                                                                         & \multirow{-2}{*}{BCE}                         & \multirow{-2}{*}{124.44M}                         & \multirow{-2}{*}{12}                                           & \multirow{-2}{*}{12}                                              & 2                                                              & 64.47                         & 20.31                         & 43.3                          & 27.93                         & 28.2                          & 36.84                         & \multirow{-2}{*}{37.16}                                  \\ \cline{2-13} 
\multicolumn{1}{c|}{}                                                                         & ZTT                                            & 128.91M                                           & 12                                                             & 10                                                                & 2                                                              & 65.23                         & 20.05                         & 44.61                         & 28.88                         & 28.17                         & 37.39                         & 37.39                                                    \\ \cline{2-13} 
\multicolumn{1}{c|}{}                                                                         & \cellcolor[HTML]{C0C0C0}                      & \cellcolor[HTML]{C0C0C0}                          & \cellcolor[HTML]{C0C0C0}                                       & \cellcolor[HTML]{C0C0C0}                                          & \cellcolor[HTML]{C0C0C0}1                                      & \cellcolor[HTML]{C0C0C0}65.51 & \cellcolor[HTML]{C0C0C0}20.65 & \cellcolor[HTML]{C0C0C0}44.23 & \cellcolor[HTML]{C0C0C0}26.45 & \cellcolor[HTML]{C0C0C0}28.46 & \cellcolor[HTML]{C0C0C0}37.06 & \cellcolor[HTML]{C0C0C0}                                 \\
\multicolumn{1}{c|}{\multirow{-7}{*}{GPT-2}}                                                  & \multirow{-2}{*}{\cellcolor[HTML]{C0C0C0}ZTTE} & \multirow{-2}{*}{\cellcolor[HTML]{C0C0C0}128.91M} & \multirow{-2}{*}{\cellcolor[HTML]{C0C0C0}12}                   & \multirow{-2}{*}{\cellcolor[HTML]{C0C0C0}10}                      & \cellcolor[HTML]{C0C0C0}2                                      & \cellcolor[HTML]{C0C0C0}64.69 & \cellcolor[HTML]{C0C0C0}20.22 & \cellcolor[HTML]{C0C0C0}44.78 & \cellcolor[HTML]{C0C0C0}29.42 & \cellcolor[HTML]{C0C0C0}28.33 & \cellcolor[HTML]{C0C0C0}37.49 & \multirow{-2}{*}{\cellcolor[HTML]{C0C0C0}\textbf{37.28}} \\ \hline
\multicolumn{1}{c|}{}                                                                         & V                                             & 354.82M                                           & 24                                                             & 0                                                                 & -                                                              & 67.63                         & 21.5                          & 49.07                         & 37.69                         & 33.31                         & 41.84                         & 41.84                                                    \\ \cline{2-13} 
\multicolumn{1}{c|}{}                                                                         & BC                                            & 354.82M                                           & 24                                                             & 24                                                                & 2                                                              & 69.64                         & 21.84                         & 49.96                         & 39.59                         & 32.27                         & 42.66                         & 42.66                                                    \\ \cline{2-13} 
\multicolumn{1}{c|}{}                                                                         &                                               &                                                   &                                                                &                                                                   & 1                                                              & 69.64                         & 23.12                         & 50.8                          & 39.83                         & 32.34                         & 43.15                         &                                                          \\
\multicolumn{1}{c|}{}                                                                         & \multirow{-2}{*}{BCE}                         & \multirow{-2}{*}{354.82M}                         & \multirow{-2}{*}{24}                                           & \multirow{-2}{*}{24}                                              & 2                                                              & 68.1                          & 22.53                         & 48.58                         & 38                            & 32.21                         & 41.88                         & \multirow{-2}{*}{42.52}                                  \\ \cline{2-13} 
\multicolumn{1}{c|}{}                                                                         & ZTT                                            & 370.67M                                           & 24                                                             & 22                                                                & 2                                                              & 69.53                         & 22.96                         & 49.49                         & 39.83                         & 32.46                         & 42.85                         & 42.85                                                    \\ \cline{2-13} 
\multicolumn{1}{c|}{}                                                                         & \cellcolor[HTML]{C0C0C0}                      & \cellcolor[HTML]{C0C0C0}                          & \cellcolor[HTML]{C0C0C0}                                       & \cellcolor[HTML]{C0C0C0}                                          & \cellcolor[HTML]{C0C0C0}1                                      & \cellcolor[HTML]{C0C0C0}69.15 & \cellcolor[HTML]{C0C0C0}22.44 & \cellcolor[HTML]{C0C0C0}50.84 & \cellcolor[HTML]{C0C0C0}39.2  & \cellcolor[HTML]{C0C0C0}32.32 & \cellcolor[HTML]{C0C0C0}42.79 & \cellcolor[HTML]{C0C0C0}                                 \\
\multicolumn{1}{c|}{\multirow{-7}{*}{\begin{tabular}[c]{@{}c@{}}GPT-2\\ Medium\end{tabular}}} & \multirow{-2}{*}{\cellcolor[HTML]{C0C0C0}ZTTE} & \multirow{-2}{*}{\cellcolor[HTML]{C0C0C0}370.67M} & \multirow{-2}{*}{\cellcolor[HTML]{C0C0C0}24}                   & \multirow{-2}{*}{\cellcolor[HTML]{C0C0C0}22}                      & \cellcolor[HTML]{C0C0C0}2                                      & \cellcolor[HTML]{C0C0C0}68.08 & \cellcolor[HTML]{C0C0C0}22.5  & \cellcolor[HTML]{C0C0C0}50.24 & \cellcolor[HTML]{C0C0C0}38.77 & \cellcolor[HTML]{C0C0C0}32.14 & \cellcolor[HTML]{C0C0C0}42.35 & \multirow{-2}{*}{\cellcolor[HTML]{C0C0C0}\textbf{42.57}} \\ \hline
\multicolumn{1}{c|}{}                                                                         & V                                             & 774.03M                                           & 36                                                             & 0                                                                 & -                                                              & 70.35                         & 21.67                         & 49.07                         & 40.4                          & 36.4                          & 43.58                         & 43.58                                                    \\ \cline{2-13} 
\multicolumn{1}{c|}{}                                                                         & BC                                            & 774.03M                                           & 36                                                             & 36                                                                & 2                                                              & 70.62                         & 25.91                         & 51.52                         & 43.99                         & 35.78                         & 45.56                         & 45.56                                                    \\ \cline{2-13} 
\multicolumn{1}{c|}{}                                                                         &                                               &                                                   &                                                                &                                                                   & 1                                                              & 71.38                         & 25.51                         & 50.8                          & 43.92                         & 35.78                         & 45.48                         &                                                          \\
\multicolumn{1}{c|}{}                                                                         & \multirow{-2}{*}{BCE}                         & \multirow{-2}{*}{774.03M}                         & \multirow{-2}{*}{36}                                           & \multirow{-2}{*}{36}                                              & 2                                                              & 71.16                         & 25.43                         & 50.55                         & 41.96                         & 35.98                         & 45.02                         & \multirow{-2}{*}{45.25}                                  \\ \cline{2-13} 
\multicolumn{1}{c|}{}                                                                         & ZTT                                            & 811.12M                                           & 36                                                             & 34                                                                & 2                                                              & 71.04                         & 25.57                         & 51.35                         & 43.95                         & 35.93                         & 45.57                         & 45.57                                                    \\ \cline{2-13} 
\multicolumn{1}{c|}{}                                                                         & \cellcolor[HTML]{C0C0C0}                      & \cellcolor[HTML]{C0C0C0}                          & \cellcolor[HTML]{C0C0C0}                                       & \cellcolor[HTML]{C0C0C0}                                          & \cellcolor[HTML]{C0C0C0}1                                      & \cellcolor[HTML]{C0C0C0}70.84 & \cellcolor[HTML]{C0C0C0}25.89 & \cellcolor[HTML]{C0C0C0}51.47 & \cellcolor[HTML]{C0C0C0}43.79 & \cellcolor[HTML]{C0C0C0}35.5  & \cellcolor[HTML]{C0C0C0}45.5  & \cellcolor[HTML]{C0C0C0}                                 \\
\multicolumn{1}{c|}{\multirow{-7}{*}{\begin{tabular}[c]{@{}c@{}}GPT-2\\ Large\end{tabular}}}  & \multirow{-2}{*}{\cellcolor[HTML]{C0C0C0}ZTTE} & \multirow{-2}{*}{\cellcolor[HTML]{C0C0C0}811.12M} & \multirow{-2}{*}{\cellcolor[HTML]{C0C0C0}36}                   & \multirow{-2}{*}{\cellcolor[HTML]{C0C0C0}34}                      & \cellcolor[HTML]{C0C0C0}2                                      & \cellcolor[HTML]{C0C0C0}70.48 & \cellcolor[HTML]{C0C0C0}25.74 & \cellcolor[HTML]{C0C0C0}51.33 & \cellcolor[HTML]{C0C0C0}43.65 & \cellcolor[HTML]{C0C0C0}35.2  & \cellcolor[HTML]{C0C0C0}45.28 & \multirow{-2}{*}{\cellcolor[HTML]{C0C0C0}\textbf{45.39}} \\ \hline
\end{tabular}
 \vspace{-0.5cm}
\end{table*}

We further validate our method on large, pre-trained checkpoints: GPT-2~\cite{radford2019language} and  OPT~\cite{zhang2023opt}. Table~\ref{tab:pertrain} reports the performance of various cycling strategies after fine-tuning on the same downstream tasks.

Across all model scales, {cycling-based methods} consistently outperform the {Vanilla } baseline. Basic Cycling  provides noticeable accuracy gains over Vanilla, demonstrating the effectiveness of reusing parameters through repeated computation. However, when early exit is applied ({BCE}), performance occasionally drops slightly due to the additional overhead introduced by optimizing intermediate outputs.

Among all approaches, {Zero-Token Tansformer (ZTT)} achieves the highest accuracy, surpassing both BC and V. The improvements indicate that incorporating a Zero Token during fine-tuning enables the model to effectively leverage repeated reasoning under a fixed parameter budget. Furthermore, the early-exit variant, {Zero-Token Transformer with Early Exit (ZTTE)}, maintains comparable accuracy to full ZT while significantly reducing computational costs. This confirms that adaptive inference can successfully scale to large pre-trained models.

Notably, as model sizes increase—such as GPT-2 Large with 811M parameters—both {ZT} and {ZTE} continue to provide strong accuracy gains while maintaining parameter efficiency. These results demonstrate the broad applicability and scalability of our proposed Zero-Token approach, making it a robust fine-tuning strategy for large-scale language models.

\subsection{Ablation Study}
\label{sec:ablation}

\begin{table}[ht]
\centering
\caption{
Results of the ablation study, where Gate represents the gating unit in the FNN, and ZT stands for Zero Token.
}
\label{tab:ab}
\resizebox{0.5\textwidth}{!}{ 
\begin{tabular}{cc|ccccc|c}
\hline

\textbf{Gate} & \textbf{ZT} & \textbf{PQ} & \textbf{ARC-c} & \textbf{ARC-e} & \textbf{LD} & \textbf{HS} & \textbf{Avg} \\ \hline
\checkmark & \checkmark & 63.55 & 17.88 & 40.39 & 15.3 & 26.84 & \textbf{32.79} \\
\checkmark & $\times$ & 62.95 & 18.04 & 39.62 & 15.77 & 26.75 & 32.63 \\
$\times$ & \checkmark & 63.25 & 18.12 & 40.07 & 15.22 & 26.7 & 32.67 \\
$\times$ & $\times$ & 63.56 & 18.41 & 40.03 & 14.42 & 26.67 & 32.62 \\ \hline
\end{tabular}
}
  \vspace{-0.8cm}
\end{table}

To pinpoint the contribution of each component, we conduct an ablation study by selectively removing the {Zero Token (ZT)} or the {Gate} in the FNN layer. The results, summarized in Table~\ref{tab:ab} (placeholder), highlight the individual and combined effects of these components.

The full model ({ZT + Gate}) achieves the highest average accuracy of {32.79\%}, demonstrating the complementary benefits of these two mechanisms. When the {Gate} is removed, the model experiences a slight performance drop, indicating that the gating mechanism refines the computation flow within the feed-forward network. Similarly, removing only the {Zero Token} leads to a comparable decrease in accuracy, suggesting that the Zero Token mechanism is crucial for \textit{dynamic cycle awareness}. 
Furthermore, when both components are disabled, the model reaches its lowest performance, confirming that these mechanisms play an essential role in optimizing reasoning efficiency and predictive accuracy. These findings reinforce that the combination of {Zero Token and Gate} provides the best trade-off between computational efficiency and performance.




\section{Conclusion}



We have presented {Zero-Token Transformer}, a parameter-sharing strategy for Transformers that comprehensively addresses the core questions of {which} layers to reuse, {how} to manage shared parameters, and {when} to stop iterating. By decoupling head and tail layers from the cyclic process and introducing a learnable Zero Token in each attention block, our approach enables {adaptive computation}, dynamically adjusting the number of reasoning steps based on the model’s confidence.
Our experiments show that this method is effective for both {training from scratch} and {fine-tuning pre-trained models}, consistently improving performance without increasing the overall parameter budget. The Zero Token mechanism not only facilitates parameter-efficient reasoning but also provides a straightforward criterion for early exiting, thereby reducing redundant computation while preserving accuracy.

These findings highlight the potential of {dynamic parameter-sharing strategies} in large-scale language models, particularly in resource-constrained scenarios. We believe that further exploration of zero-token prompts, gating mechanisms, and cyclic architectures will lead to increasingly efficient and adaptive Transformer-based designs in the future.




\newpage
\appendix
\onecolumn
\section{Experimental Setup}
\label{es}

\subsection{Evaluation Details}

To assess the effectiveness of our proposed method, we evaluate models on a diverse set of well-established NLP benchmarks. These benchmarks span four key reasoning tasks: \textbf{commonsense physical reasoning, multiple-choice question answering, long-term context recall, and natural language inference}. For all datasets, we report \textbf{accuracy (ACC)} as the primary evaluation metric.

\subsubsection{Reasoning: PIQA}
\textbf{Dataset:} The \textit{Physical Interaction Question Answering (PIQA)} dataset~\cite{bisk2020piqa} evaluates a model’s ability to reason about \textbf{everyday physical interactions}. It consists of multiple-choice questions that require an understanding of how objects and tools function in real-world scenarios.

\textbf{Task Objective:} Given a short natural language query, the model must select the most plausible solution from two candidate answers. This task assesses the model's ability to infer \textbf{practical physical knowledge} beyond simple memorization.

\textbf{Evaluation Metric:} Accuracy (ACC), measuring the proportion of correctly predicted answers.

\subsubsection{Multiple-Choice Question Answering: ARC Challenge and ARC Easy}
\textbf{Dataset:} The \textit{AI2 Reasoning Challenge (ARC)}~\cite{clark2018think} is a standardized multiple-choice QA benchmark designed to evaluate a model’s ability to answer \textbf{science-related questions}. It consists of two subsets:
\begin{itemize}
    \item \textbf{ARC Challenge:} A more difficult subset requiring complex reasoning and deeper knowledge retrieval.
    \item \textbf{ARC Easy:} A simpler subset containing factual questions that can often be answered with surface-level understanding.
\end{itemize}

\textbf{Task Objective:} The model is provided with a science-related question and four answer choices, from which it must select the correct one. The dataset requires a combination of \textbf{commonsense reasoning, logical inference, and scientific knowledge} to achieve high performance.

\textbf{Evaluation Metric:} Accuracy (ACC), computed as the percentage of correctly answered questions.

\subsubsection{Long-Term Context Recall: LAMBADA}
\textbf{Dataset:} The \textit{LAMBADA} dataset~\cite{paperno2016lambada} is specifically designed to assess a model’s capability for \textbf{long-range context comprehension}. Unlike standard language modeling tasks, LAMBADA requires a model to retain and process information over an extended passage to predict a crucial missing word.

\textbf{Task Objective:} Given a \textbf{long contextual passage}, the model must predict the  final missing word  of the last sentence. The difficulty arises from the fact that the target word is nearly impossible to guess without understanding the full passage.

\textbf{Evaluation Metric:} Accuracy (ACC), where a prediction is considered correct if the \textbf{entire target word} matches the ground truth exactly.

\subsubsection{Natural Language Inference: HellaSwag}
\textbf{Dataset:} The \textit{HellaSwag} dataset~\cite{zellers2019hellaswag} is an advanced benchmark designed to evaluate \textbf{commonsense inference and story continuation}. It builds on the SWAG dataset by incorporating adversarial filtering, making it more challenging for models to rely on surface-level heuristics.

\textbf{Task Objective:} Given an  incomplete story or event description , the model must select the most  logical next step  from four possible continuations. This requires strong  contextual understanding  and the ability to anticipate  real-world event progressions .

\textbf{Evaluation Metric:} Accuracy (ACC), measuring how often the model correctly predicts the most plausible continuation.

\subsection{Training and Fine-Tuning Settings}

In this section, we describe the training settings for both \textbf{pre-training from scratch} and \textbf{fine-tuning of pre-trained models}. The \textbf{fine-tuning stage} is required for all models before final evaluation, while models trained from scratch undergo \textbf{both pre-training and fine-tuning}. The fine-tuning hyperparameters are kept consistent across both settings.

\subsubsection{Pre-Training from Scratch}

For models trained from scratch, we first conduct pre-training on the \textit{C4 English dataset}~\cite{raffel2020exploring}. The pre-training process follows these configurations:

\paragraph{Pre-Training Protocol}
\begin{itemize}
    \item \textbf{Dataset:} The \textit{C4 (Colossal Clean Crawled Corpus)} English subset.
    \item \textbf{Computing Resources:} We utilize an \texttt{A800 GPU cluster} for training.
    \item \textbf{Batch Size per GPU:} 80, with \textbf{gradient accumulation} to maintain a global batch size of 256.
    \item \textbf{Training Steps:} The model is trained for a total of \textbf{10B tokens}.
    \item \textbf{Optimizer:} AdamW~\cite{loshchilov2018decoupled} with a weight decay of 0.01.
    \item \textbf{Learning Rate:} A linear warmup is applied for the first 1\% of total steps, followed by a cosine decay schedule.
    \item \textbf{Precision:} Training is performed in \textbf{half-precision (FP16)} to optimize memory efficiency.
\end{itemize}

After pre-training, the models proceed to the \textbf{fine-tuning} stage before being evaluated on downstream tasks.

\subsubsection{Fine-Tuning Settings}

Before evaluating on the test datasets, we fine-tune our models using the corresponding training sets. \textbf{For pre-trained models, only fine-tuning is performed}, while models trained from scratch undergo \textbf{both pre-training and fine-tuning}. The fine-tuning process is conducted under the same computational settings as pre-training.

\paragraph{Fine-Tuning Protocol}
\begin{itemize}
    \item \textbf{Fine-Tuning Epochs:} Each dataset is fine-tuned for \textbf{3 epochs}.
    \item \textbf{Batch Size per GPU:} 20, with \textbf{gradient accumulation} ensuring an effective batch size of 80.
    \item \textbf{Optimizer:} AdamW with a 0.01 weight decay.
    \item \textbf{Learning Rate:} The default Hugging Face \texttt{Trainer} API learning rate is used.
    \item \textbf{Prompt Engineering:} We utilize prompt templates from \texttt{promptsource} to better adapt models to the task format.
    \item \textbf{Computing Resources:} The same \texttt{A800 GPU cluster} is used as in pre-training.
    \item \textbf{Training Framework:} Fine-tuning is implemented with Hugging Face’s \texttt{Trainer API}.
\end{itemize}

\paragraph{Dataset-Specific Fine-Tuning Details}
Fine-tuning is performed on the following datasets before model evaluation. The details of each dataset, including the number of training examples, are presented in \cref{tab:finetune}.

\begin{table}[h]
\centering
\small
\caption{Fine-tuning settings for each dataset, including the number of training epochs and dataset sizes.}
\label{tab:finetune}
\begin{tabular}{l|c|c|c}
\hline
\textbf{Dataset} & \textbf{Epochs} & \textbf{Training Size} & \textbf{Validation Size} \\ \hline
PIQA            & 3               & 16,000                 & 1,838                     \\
ARC Challenge   & 3               & 1,119                  & 1,172                     \\
ARC Easy        & 3               & 2,251                  & 2,376                     \\
LAMBADA         & 3               & 4,869                  & 4,869                     \\
HellaSwag       & 3               & 39,905                 & 10,042                    \\ \hline
\end{tabular}
\end{table}

\subsection{Early-Exit Training Settings}

To ensure \textbf{effective intermediate predictions} when early-exit mechanisms are applied, we implement \textbf{additional training for intermediate classifier heads}. This helps maintain meaningful intermediate outputs, preventing degradation in performance due to premature exits.

\subsubsection{Classifier Placement}
\begin{itemize}
    \item \textbf{Simple Cycling:} The classification head is placed \textbf{only at the final output layer}.
    \item \textbf{Head-Tail Separation:} The classification head is placed at \textbf{both the final layer and the last shared layer before cycling begins}.
\end{itemize}

\subsubsection{Training Strategy for Early-Exit Models}

To optimize models for early exits, we introduce additional supervision at intermediate layers. Instead of relying solely on the final output, we ensure that \textbf{multiple exit points} are trained effectively.

\begin{itemize}
    \item \textbf{Intermediate Supervision:} The model is trained to produce meaningful predictions at designated early-exit points.
    \item \textbf{Exit Point Optimization:} Models with \textbf{multiple cycling blocks} undergo training to align their intermediate outputs with final predictions, improving robustness across different exit depths.
    \item \textbf{Gradual Refinement:} The early-exit heads are optimized using the same fine-tuning data, ensuring consistency across all prediction layers.
\end{itemize}

By integrating these early-exit classifiers and \textbf{fine-tuning them separately}, we ensure that models can \textbf{gracefully exit at earlier layers without sacrificing predictive accuracy}. This design allows our method to maintain efficiency while preserving strong performance across different computational budgets.

\section{More Experimental Results}
\label{sec:more_experiments}

To further analyze the effectiveness of our method, we present additional  adaptive reasoning loop  and ablation experiments in \cref{tab:ad_detail} and \cref{tab:adaptive_detail}.

\subsection{Analysis of Adaptive Reasoning Loops}
\begin{table}[]
\centering
\small  
\caption{
More detailed results on adaptive reasoning loop counts.
}

\label{tab:adaptive_detail}
\begin{tabular}{cc|ccccc|c}
\hline
\rowcolor[HTML]{C0C0C0} 
\multicolumn{2}{c|}{\cellcolor[HTML]{C0C0C0}\textbf{p}}                           & \textbf{PQ}                   & \textbf{ARC-c}                & \textbf{ARC-e}                & \textbf{LD}                   & \textbf{HS}                   & \textbf{Avg}                  \\ \hline
\multicolumn{1}{c|}{}                               & Loop                        & 1.82                          & 1.61                          & 1.42                          & 1.72                          & 1.33                          & 1.58                          \\
\multicolumn{1}{c|}{\multirow{-2}{*}{\textbf{0.2}}} & \cellcolor[HTML]{FFCE93}Acc & \cellcolor[HTML]{FFCE93}63.02 & \cellcolor[HTML]{FFCE93}17.99 & \cellcolor[HTML]{FFCE93}39.04 & \cellcolor[HTML]{FFCE93}15.26 & \cellcolor[HTML]{FFCE93}26.79 & \cellcolor[HTML]{FFCE93}32.42 \\ \hline
\multicolumn{1}{c|}{}                               & Loop                        & 2.9                           & 3.23                          & 3.13                          & 3.56                          & 3.77                          & 3.32                          \\
\multicolumn{1}{c|}{\multirow{-2}{*}{\textbf{0.5}}} & \cellcolor[HTML]{FFCE93}Acc & \cellcolor[HTML]{FFCE93}63.22 & \cellcolor[HTML]{FFCE93}18.17 & \cellcolor[HTML]{FFCE93}41.46 & \cellcolor[HTML]{FFCE93}15.27 & \cellcolor[HTML]{FFCE93}26.87 & \cellcolor[HTML]{FFCE93}33    \\ \hline
\multicolumn{1}{c|}{}                               & Loop                        & 3.87                          & 3.42                          & 3.47                          & 3.93                          & 3.99                          & 3.74                          \\
\multicolumn{1}{c|}{\multirow{-2}{*}{\textbf{0.7}}} & \cellcolor[HTML]{FFCE93}Acc & \cellcolor[HTML]{FFCE93}64.15 & \cellcolor[HTML]{FFCE93}18.34 & \cellcolor[HTML]{FFCE93}41.25 & \cellcolor[HTML]{FFCE93}14.67 & \cellcolor[HTML]{FFCE93}26.97 & \cellcolor[HTML]{FFCE93}33.08 \\ \hline
\multicolumn{1}{c|}{}                               & Loop                        & 4                             & 4                             & 44                            & 4                             & 4                             & 4                             \\
\multicolumn{1}{c|}{\multirow{-2}{*}{\textbf{1}}}   & \cellcolor[HTML]{FFCE93}Acc & \cellcolor[HTML]{FFCE93}63.55 & \cellcolor[HTML]{FFCE93}18    & \cellcolor[HTML]{FFCE93}41.04 & \cellcolor[HTML]{FFCE93}13.59 & \cellcolor[HTML]{FFCE93}26.93 & \cellcolor[HTML]{FFCE93}32.62 \\ \hline
\end{tabular}
 \vspace{-0.5cm}
\end{table}
\cref{tab:adaptive_detail} presents results on our \textbf{adaptive reasoning loop mechanism}, where the model dynamically determines the number of iterations based on the \textbf{Zero Attention threshold ($P$)}.

\textbf{Key observations:}
\begin{itemize}
    \item \textbf{Low threshold ($P = 0.2$)} results in early exits (1.58 cycles) but slightly lower accuracy (\textbf{32.42\%}).
    \item \textbf{Balanced performance at $P = 0.5$}: The model averages \textbf{3.31 cycles} and reaches \textbf{33.00\% accuracy}, achieving strong efficiency gains.
    \item \textbf{Higher thresholds ($P = 0.7$)} lead to more computation (3.74 cycles) and slight accuracy gains (\textbf{33.08\%}), but with diminishing returns.
    \item \textbf{Full computation ($P = 1$)} does not significantly outperform adaptive strategies, confirming that early exit can maintain performance.
\end{itemize}

These results demonstrate that adaptive early exit strategies reduce computation while maintaining accuracy, with \textbf{$P = 0.5$} being the most efficient trade-off.

\subsection{Ablation Study on Zero Token and Gating Mechanism}
\begin{table}[]
\centering
\small  
\caption{
More detailed ablation study results, including the detailed outcomes of each early exit.
}


\label{tab:ad_detail}
\begin{tabular}{c|c|c|c|c|ccccc|c|c}
\hline
\textbf{Models}          & \textbf{Size}           & \textbf{\begin{tabular}[c]{@{}c@{}}All \\ Layers\end{tabular}} & \textbf{\begin{tabular}[c]{@{}c@{}}Looped \\ Layers\end{tabular}} & \textbf{\begin{tabular}[c]{@{}c@{}}Loop \\ Count\end{tabular}} & \textbf{PQ} & \textbf{ARC-c} & \textbf{ARC-e} & \textbf{LD} & \textbf{HS} & \textbf{Avg} & \textbf{Model\_Avg}    \\ \hline
\multirow{4}{*}{Wo Gate} & \multirow{4}{*}{60.66M} & \multirow{4}{*}{3}                                             & \multirow{4}{*}{1}                                                & 1                                                              & 63.06       & 18.69          & 40.03          & 16.13       & 26.64       & 32.91        & \multirow{4}{*}{32.68} \\
                         &                         &                                                                &                                                                   & 2                                                              & 63.6        & 17.83          & 40.36          & 15.34       & 26.76       & 32.78        &                        \\
                         &                         &                                                                &                                                                   & 3                                                              & 63.11       & 18.03          & 39.86          & 15.71       & 26.84       & 32.71        &                        \\
                         &                         &                                                                &                                                                   & 4                                                              & 63.22       & 17.92          & 40.03          & 13.7        & 26.65       & 32.3         &                        \\ \hline
\multirow{4}{*}{Wo ZT}   & \multirow{4}{*}{61.76M} & \multirow{4}{*}{3}                                             & \multirow{4}{*}{1}                                                & 1                                                              & 62.72       & 17.75          & 39.81          & 15.37       & 26.81       & 32.49        & \multirow{4}{*}{32.63} \\
                         &                         &                                                                &                                                                   & 2                                                              & 62.92       & 18.22          & 40.32          & 16.5        & 26.76       & 32.94        &                        \\
                         &                         &                                                                &                                                                   & 3                                                              & 63.02       & 17.77          & 40.19          & 16.22       & 26.88       & 32.82        &                        \\
                         &                         &                                                                &                                                                   & 4                                                              & 63.12       & 18.43          & 38.17          & 14.98       & 26.55       & 32.25        &                        \\ \hline
\end{tabular}
 \vspace{-0.5cm}
\end{table}
\cref{tab:ad_detail} provides a detailed breakdown of our ablation study, evaluating the impact of the \textbf{Zero Token (ZT)} mechanism and the \textbf{gating unit (Gate)} in the feed-forward network (FNN). The results highlight the individual and combined contributions of these components.

\textbf{Key findings:}
\begin{itemize}
    \item The full model (\textbf{ZT + Gate}) achieves the highest accuracy (\textbf{32.79\%}), demonstrating that both components are essential.
    \item Removing the \textbf{Gate} leads to a slight performance drop (\textbf{32.68\%}), suggesting that gating helps refine reasoning.
    \item Removing the \textbf{Zero Token} reduces accuracy further (\textbf{32.63\%}), indicating its role in guiding iterative reasoning.
    \item The baseline model (without ZT and Gate) achieves the lowest accuracy (\textbf{32.62\%}), confirming that both components contribute positively.
\end{itemize}

These results validate that both \textbf{Zero Token and Gate} are essential for maximizing model efficiency and reasoning quality.

\end{document}